\documentclass[runningheads]{llncs}
\usepackage{graphicx}

\usepackage{tikz}
\usepackage{comment}
\usepackage{amsmath,amssymb} 
\usepackage{color}

\usepackage[accsupp]{axessibility}  

\usepackage{url}            
\usepackage{booktabs}       
\usepackage{amsfonts}       
\usepackage{nicefrac}       
\usepackage{microtype}      
\usepackage{xcolor}         

\usepackage{epsfig}
\usepackage{graphicx}
\usepackage{amsmath}
\usepackage{amssymb}

\usepackage{mathtools}
\usepackage{multirow}

\usepackage{float}
\usepackage{wrapfig}
\usepackage{soul}

\usepackage{caption}
\usepackage{subcaption}

\usepackage[pagebackref,breaklinks,colorlinks]{hyperref}
\usepackage{orcidlink}

\usepackage[width=122mm,left=12mm,paperwidth=146mm,height=193mm,top=12mm,paperheight=217mm]{geometry}

\usepackage[capitalize]{cleveref}
\crefname{section}{Sec.}{Secs.}
\Crefname{section}{Section}{Sections}
\Crefname{table}{Table}{Tables}
\crefname{table}{Tab.}{Tabs.}

\begin{document}

\pagestyle{headings}
\mainmatter
\def\ECCVSubNumber{5591}  

\title{Semi-Supervised Learning of Optical Flow by Flow Supervisor} 

\titlerunning{Semi-Supervised Learning of Optical Flow by Flow Supervisor}
%
\author{Woobin Im \and
Sebin Lee \and
Sung-Eui Yoon}

\authorrunning{W. Im et al.}
%
\institute{Korea Advanced Institute of Science and Technology (KAIST),\\
\email{\{iwbn,seb.lee,sungeui\}@kaist.ac.kr}}

\maketitle

\begin{abstract}
   A training pipeline for optical flow CNNs 
consists of a pretraining stage on a synthetic dataset
followed by a fine tuning stage on a target dataset.
However, obtaining ground truth flows from a target video 
requires a tremendous effort.
This paper proposes a practical fine tuning method to adapt a pretrained model to a target dataset without ground truth flows, which has not been explored extensively.
Specifically, 
we propose a flow supervisor for self-supervision, which consists of parameter separation and a student output connection.
This design is aimed at
stable convergence and better accuracy over
conventional self-supervision methods which are unstable on the fine tuning task.
Experimental results show the effectiveness of our method
compared to different self-supervision methods for semi-supervised learning.
In addition, we achieve meaningful improvements 
over state-of-the-art optical flow models
on Sintel and KITTI benchmarks
by exploiting additional unlabeled datasets.
Code is available at \url{https://github.com/iwbn/flow-supervisor}.

\end{abstract}

\section{Introduction}

Optical flow 
describes the pixel-level displacement in two images,
and is a fundamental step for various motion understanding tasks in computer vision.
Recently, supervised deep learning methods have shown
remarkable performance in terms of
overcoming challenges -- such as motion blur, change of brightness and color, deformation, and occlusion --
and predicting more accurate flows.
The key to success is end-to-end learning from large-scale data.
For optical flow learning,
large-scale datasets have been released~\cite{dosovitskiy2015flownet,ilg2017flownet,Butler:ECCV:2012} and
deep architectures have been advanced~\cite{dosovitskiy2015flownet,sun2018pwc,teed2020raft}.

Building a good optical flow model on a target dataset
is critical;
most training data are synthetic, and it requires tremendous effort
to label random video frames by pixel-wise dense correspondences.
Thus, to obtain a good model on a target dataset,
synthetic training set generation~\cite{sun2021autoflow,mayer2018makes} and
GAN-based adaptation~\cite{lai2017semi,yan2020optical} have been studied. 
In addition, unsupervised loss functions~\cite{meister2018unflow,stone2021smurf} 
-- used without
ground truth -- have been proposed.
However, generating a synthetic dataset for a target domain
is often computationally expensive or confined to a specific domain.
Moreover,
unsupervised losses do not reach the state-of-the-arts, compared to supervised methods.
Therefore, there have been needs for a simpler, general, and high-performance
method to build a better model on a target dataset.

\begin{wrapfigure}{r}{0.5\textwidth}
\centering
 \includegraphics[trim={0mm 33.8mm 117mm 40.6mm},clip,width=1.0\linewidth]{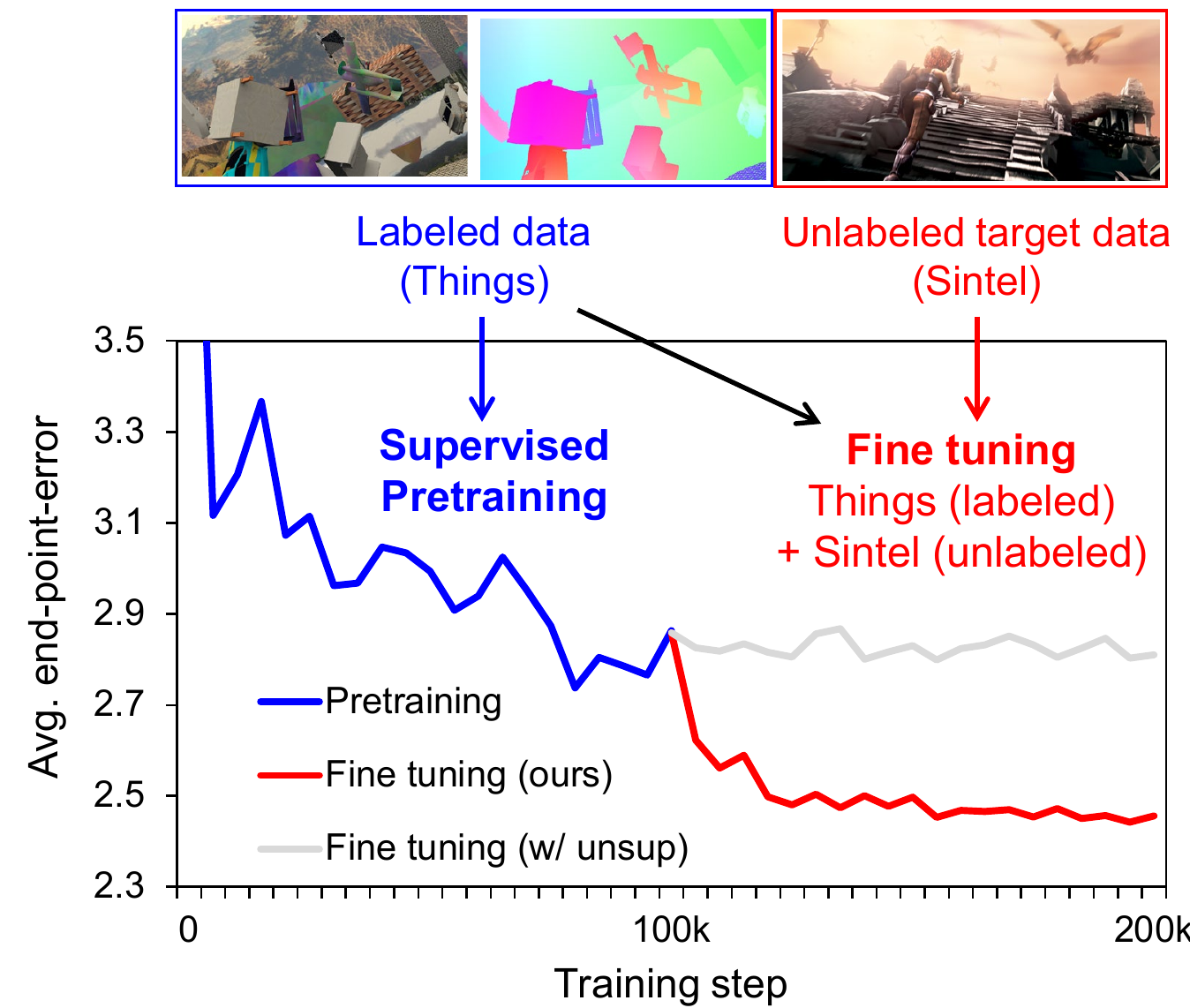}
\caption{\textbf{End-point-error on Sintel}. Our method is used to adapt a pretrained model
to a target domain without a target domain label;
it is designed to overcome an unstable convergence and 
low accuracy in traditional methods.
For instance, our method outperforms the unsupervised loss (Eq.~\ref{eq:unsuploss}) in fine tuning, which makes our method favorably better
}
\label{fig:hero}
\end{wrapfigure}

In this paper, 
we propose a fine tuning strategy for semi-supervised optical flow learning,
which helps to build a better model on unlabeled or partly-labeled target datasets.
Fig.~\ref{fig:hero} demonstrates the concept of our approach.
Our method is a fine tuning method, 
where the pretrained network is adapted to the unlabeled target data.
In the fine tuning stage, we use a labeled dataset with an unlabeled target dataset,
further reducing errors on the target dataset.

To build our method, we investigate 
unsupervised and self-supervised approaches,
where a network learns optical flow by unlabeled samples.
However, the existing unsupervised loss does not show 
better performance in the fine tuning stage (Fig.~\ref{fig:hero}).
Moreover, self-supervision methods tend to show highly unstable behavior or lower performance in the fine tuning stage.
To address the issue, 
we propose our flow supervisor
with two strategies: the parameter separation 
and passing student outputs,
which are effective
for higher performance and stable semi-supervised learning.
As shown in Fig.~\ref{fig:hero}, our fine tuning method
clearly reduces the error of the pretrained model, even without 
a label of the target dataset.

To summarize, 
we propose a semi-supervised fine tuning strategy to improve an optical flow network on a target dataset, which has not been explored extensively.
Our strategy is distinguished by the flow supervisor module, designed with
the parameter separation and 
passing student outputs.
We show the effectiveness of our method by comparing it with
alternative self-supervision methods, and
confirm that our approach stabilizes the learning process
and results in better accuracy.
In addition,
we test our method on Sintel and KITTI benchmarks
and achieve meaningful improvements over the state-of-the-arts by 
exploiting additional unlabeled data.

\section{Related Work}
\noindent \textbf{Supervised optical flow}
has been studied with the development of 
datasets for optical flow learning~\cite{dosovitskiy2015flownet,ilg2017flownet,sun2021autoflow}
and
the advances of deep network architectures~\cite{dosovitskiy2015flownet,sun2018pwc,teed2020raft}.
Due to the high annotation cost and label ambiguity in a raw video, 
synthetic datasets have been made, where 
optical flow fields are generated together with images~\cite{mayer2018makes,dosovitskiy2015flownet,ilg2017flownet,sun2021autoflow}.
Along with the datasets, network architectures for optical flow have been
significantly improved by the cost-volume~\cite{dosovitskiy2015flownet,sun2018pwc,yang2019volumetric}, warping~\cite{ilg2017flownet,sun2018pwc}, and refinement scheme~\cite{hur2019iterative,teed2020raft}.
Although generalization has been improved thanks to the synthetic datasets and the network architectures,
it is still difficult to achieve better performance 
while being blind to a target domain~\cite{novak2020semi}.

\noindent \textbf{Unsupervised optical flow}
is another stream of optical flow research, 
where optical flow is learned without an expensive 
labeling or label generation process~\cite{meister2018unflow,Im_2020_ECCV,jonschkowski2020matters,stone2021smurf}.
With the advanced deep architectures from supervised optical flow,
unsupervised optical flow studies have focused on designing loss functions.
Previous work has mainly focused on
the fully unsupervised setting, while
we explore semi-supervised optical flow where labeled data is accessible in addition to unlabeled target domain data.

\noindent \textbf{Semi-supervised optical flow}
has been studied to utilize unlabeled target domain data
with existing labeled synthetic data~\cite{lai2017semi}.
The experimental setting -- similar to unsupervised domain adaptation~\cite{ganin2015unsupervised} --
is designed since it is relatively easier to exploit synthetic training datasets
than annotating images of a target dataset.
A simple baseline method for a target domain would be 
using a traditional unsupervised loss~\cite{meister2018unflow}
and supervised learning, 
which, unfortunately, has shown inferior performance than the 
supervision-only training~\cite{lai2017semi,novak2020semi} (Table~\ref{tab:da_result}).
Thus, reducing the domain gap~\cite{lai2017semi} and
stabilizing unsupervised loss gradients~\cite{novak2020semi} have been proposed.

In this work, we introduce the flow supervisor,
which consists of the separate parameters and the student output connection.
We found our design scheme is superior to the baseline designs
in terms of training stability and performance 
in the semi-supervised setting.

\noindent \textbf{Knowledge distillation and self-supervision} in neural networks
are proposed to train a network under the guidance of a teacher network~\cite{hinton2015distilling,stanton2021does} or itself~\cite{chen2021exploring,grill2020bootstrap}.
Interestingly, the technique can build 
a better student network even when the same network architecture is used for both student and teacher, i.e., self-distillation~\cite{hinton2015distilling,stanton2021does}.
In the optical flow field, knowledge distillation and
self-supervision have been studied 
actively in the context of unsupervised optical flow~\cite{liu2021learning,liu2019selflow,stone2021smurf}.
Generally, these methods can be interpreted as learning using privileged information~\cite{vapnik2015learning}, in that 
a student usually sees a limited view, e.g., cropped images,
while a teacher is given a privileged view, e.g., full images.
In this work, we investigate the effectiveness of 
self-supervision in terms of semi-supervised optical flow,
which has not been explored in previous work.
In addition, we found that the traditional self-supervision 
method for optical flow~\cite{stone2021smurf} 
tends to make a loss diverge 
in the semi-supervised setting. 

Thus, we propose a novel self-supervision method 
to ameliorate the unstable convergence;
we found that the parameter separation of a teacher model 
and passing student output 
are the key components to make the training stable.
By applying our method, a network can successfully exploit the unlabeled target data in the semi-supervised setting. 
In addition, we show that our method can address the lack of a target labeled dataset, e.g., 200 labeled pairs for KITTI, 
by the ability to utilize unlabeled samples.

\section{Approach}

\subsection{Preliminaries on Deep Optical Flow}
Deep optical flow estimation is defined with
an optical flow estimator $f_\theta(\mathbf{x}_1, \mathbf{x}_2)$, which predicts 
optical flow $\hat{\mathbf{y}}\in \mathbb{R}^{H\times W\times 2}$ from 
two images $\mathbf{x}_1, \mathbf{x}_2 \in \mathbb{R}^{H \times W \times C}$, where
$H$ is height, $W$ is width, and $C$ is channel.

In supervised learning, 
we train $f_\theta$ by 
minimizing the supervised loss:
\begin{equation}
    \mathcal{L}_\text{sup}(\theta) = \mathbb{E}_{(\mathbf{x}_1, \mathbf{x}_2, \mathbf{y})\sim p_s}[\ell_\text{sup}(f_\theta (\mathbf{x}_1, \mathbf{x}_2), \mathbf{y})],\label{eq:suploss}
\end{equation}
where $p_s$ is the labeled data distribution,
$\mathbf{y}$ is the ground truth optical flow, and
$\ell_\text{sup}$ is $L_1$,
$L_2$~\cite{sun2018pwc} or 
the generalized Charbonnier loss~\cite{meister2018unflow}.

On the other hand, an unsupervised method
defines a loss function 
with a differentiable target function $\ell_\text{unsup}(\cdot)$
which can be computed without a ground truth $\mathbf{y}$,
resulting in the unsupervised loss:
\begin{equation}
    \mathcal{L}_\text{unsup}(\theta) = \mathbb{E}_{(\mathbf{x}_1, \mathbf{x}_2)\sim p_u}[\ell_\text{unsup}(f_\theta (\mathbf{x}_1, \mathbf{x}_2), \mathbf{x}_1, \mathbf{x}_2)],\label{eq:unsuploss}
\end{equation}
where $p_u$ is an unlabeled data distribution.
Most commonly, $\ell_\text{unsup}$ is defined with
the photometric loss $\ell_\text{photo} =  \rho(\text{warp}(\mathbf{x}_2,\hat{\mathbf{y}}) - \mathbf{x}_1)$ 
where $\rho$ is the Charbonnier loss~\cite{meister2018unflow} and $\text{warp}(\cdot)$ is the differentiable backward warping
operation~\cite{jaderberg2015spatial}.

\subsection{Problem Definition and Background}
We train our model on
labeled data and unlabeled data,
which is
similar to experimental settings that appear in \cite{lai2017semi,novak2020semi}.
For instance, FlyingThings3D (rendered scene) and KITTI (driving scene) can be considered as labeled and unlabeled datasets, respectively.
We aim to build a high-performance model on a target domain, with only a labeled synthetic dataset and an unlabeled target dataset.
This is a practical scenario since
a synthetic dataset is relatively inexpensive and publicly available,
whereas specific target domain data is rarely annotated.

We focus on designing a self-supervision method with a stable convergence and better accuracy,
since it is not trivial to adopt 
unsupervised losses and self-supervision for semi-supervised learning.
First, unsupervised loss functions 
often lead a network to a worse local minimum when it is naively fine tuned from supervised (pretrained) weights.
Second, traditional self-supervision strategies 
for semi-supervised learning 
do not converge well.
Thus, we propose a
simple and effective practice for semi-supervised learning,
where
we utilize synthetic datasets for better performance on a target dataset.
By our strategy,
a network can successfully exploit the unlabeled target data for better
fine tuning performance.
Detailed method is discussed in the next section.
\begin{figure}[t]
 \centering
\begin{subfigure}[b]{0.48\textwidth}
 \centering
 \includegraphics[trim={72mm 106.5mm 105mm 43.346mm},clip,width=\textwidth]{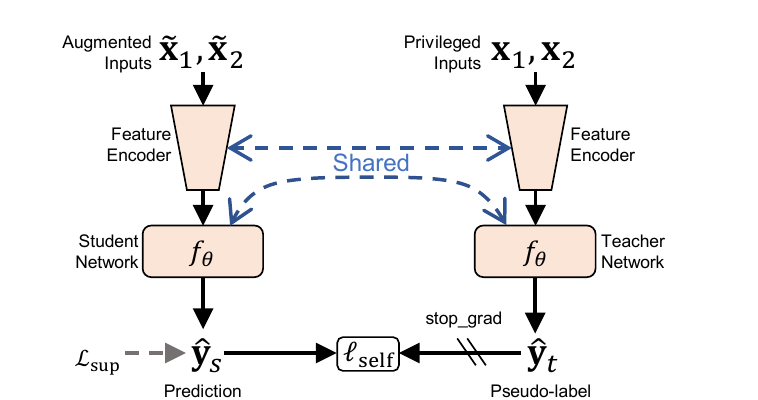}
 \caption{Self-supervision}\label{subfig:rt_concept_a}
 \end{subfigure}
 \hfill
\begin{subfigure}[b]{0.48\textwidth}
 \centering
 \includegraphics[trim={152mm 106.5mm 26mm 43.346mm},clip,width=\textwidth]{figure/rt_concept_simple.pdf}
 \caption{Flow supervisor}\label{subfig:rt_concept_b}
\end{subfigure}
\centering
\caption{
\textbf{(a)
Self-supervision} for optical flow
is configured with a teacher network which is given privileged images as
an input, i.e., full images before cropping.
\textbf{(b) Flow supervisor} 
reviews the student flow $\hat{\mathbf{y}}_s$ and outputs
the pseudo-label $\hat{\mathbf{y}}_\text{FS}$ 
to supervise the student without ground truth flows. 
We use the separate flow supervisor with parameter $\phi$, 
which improves the stability and accuracy
}
\label{fig:rt_concept}
\end{figure}

\subsection{Flow Supervisor}\label{subsec:residual_teacher}
Our design for semi-supervised optical flow learning
is based on self-supervised learning where
a student network learns from a pseudo-label predicted by a teacher network.
This concept has been explored in the optical flow field in terms of unsupervised learning, 
which is not directly applicable in the semi-supervised case due to unstable convergence.

Thus, we introduce 
two distinctive design schemes compared to the existing self-supervision techniques for optical flow, depicted in Fig.~\ref{fig:rt_concept}.
First, we introduce a supervisor parameter $\phi$, distinguished from the student parameter $\theta$, which learns to supervise the student network.
Second, we add a connection from the student network to the supervisor network,
which enables the supervisor network to learn conditional 
knowledge, i.e., $P(\mathbf{y}|\hat{\mathbf{y}}_s)$,
instead of predicting from scratch.

We define the student network $f_\theta$ and the flow supervisor network $f_\phi$,
as shown in Fig.~\ref{subfig:rt_concept_b}.
The student network $f_\theta$ includes a feature encoder and a flow decoder;
for simplicity, we abstract the feature encoder and decoder parameters
with $\theta$.
The flow decoder has the internal feature $\mathbf{h}_s$, and outputs the predicted flow $\hat{\mathbf{y}}_s$.
The student network is the optical flow network used for inference,
whereas the supervisor network is only used for training; 
this results in no additional computational cost for the testing time.
In training time, we use the flow supervisor (FS) loss function
$\mathcal{L}_\text{FS}$ to supervise the student flow $\hat{\mathbf{y}}_s$
with the teacher flow $\hat{\mathbf{y}}_\textit{FS}$:
\begin{equation}
    \mathcal{L}_\text{FS}(\theta) = \mathbb{E}_{d_s\sim p_s, d_u\sim p_u}[\ell_\text{sup}(d_s) + \alpha \ell_\text{FS}(d_u)],
\end{equation}
where $\ell_\text{FS}(\cdot) = \rho(\hat{\mathbf{y}}_\textit{FS} - \hat{\mathbf{y}}_s)$, $\alpha$ is a hyper-parameter weight and
$(d_s, d_u)$ are sampled data from $(p_s, p_u)$. 
We use loss function $\mathcal{L}_\text{FS}$ for the student network to learn from both labeled and unlabeled data.

\paragraph{Separate parameters.} 
We empirically observed that the plain self-supervision (Fig.~\ref{subfig:rt_concept_a}) leads to divergence in semi-supervised optical flow learning.
This undesirable behavior is also observed in siamese self-supervised learning, where preventing undesirable equilibria is important~\cite{grill2020bootstrap,chen2021exploring}.
As our solution,
we have the separate module, i.e., the flow supervisor to prevent
the unstable learning behavior.
Our flow supervisor is related to the predictor module in self-supervised learning work~\cite{chen2021exploring}, which also prevents
the unstable training.
We also compare our design with the mean-teacher~\cite{tarvainen2017mean,grill2020bootstrap} with exponential moving average (EMA),
and a fixed teacher~\cite{zoph2020rethinking} in Fig.~\ref{fig:da_result}, since these designs have been widely adopted in semi-supervised learning. 

\paragraph{Passing student outputs.} 
We design our supervisor model to have input nodes for
student outputs.
Thus, the teacher output flow is conditioned by the student output.
Specifically, our teacher model includes a residual function $\Delta f_\phi(\cdot)$, such that
$f_\phi(\mathbf{x}, \hat{\mathbf{y}}_s, \mathbf{h}_s) = \Delta f_\phi(\cdot) + \hat{\mathbf{y}}_s$.
We realize this concept with the residual flow decoder in the RAFT~\cite{teed2020raft} architecture.

In residual learning, it has been believed and shown that 
learning a residual function $\Delta f(\mathbf{x}) = f(\mathbf{x}) - \mathbf{x}$, 
instead of the original function $f(\mathbf{x})$, is better for 
deeper inference~\cite{he2016deep} or domain discrepancy modeling~\cite{long2016unsupervised}.
With residual teacher function $\Delta f_\phi$, the flow supervisor loss is 
reformulated as:
\begin{equation}
    \ell_\text{FS} = \rho(\hat{\mathbf{y}}_\textit{FS} - \hat{\mathbf{y}}_s) = \rho(f_\phi(\hat{\mathbf{y}}_s) -\hat{\mathbf{y}}_s)=\rho(\Delta f_\phi(\hat{\mathbf{y}}_s)).
    \label{eq:residual}
\end{equation}

\paragraph{Relation of $f_\phi(\cdot)$ to a meta learner.}
Meta-learning~\cite{bengio1990learning,finn2017model} defines how to learn by learning an update rule $\Delta \theta^t$ within the parameter update $\theta^{t+1} \leftarrow \theta^t + \Delta \theta^t$.
The residual function $\Delta f_\phi(\dots,\hat{\mathbf{y}}_s)$ is regarded as a meta learner predicting update rule $\Delta \theta^t$,
conditioned by student predictions.
Assuming $\rho$ is the square function, 
learning by $\mathcal{L}_\text{FS}$ is equivalent to using
the update rule $\Delta \theta^t = -2\frac{\partial \hat{\mathbf{y}}_s}{\partial \theta^t}^T\cdot\Delta f_\phi(\hat{\mathbf{y}}_s)$;
where $\phi$ is the parameter of our flow supervisor.
That is, the learning rule is learned by the supervisor parameter 
$\phi$ to supervise the student parameter $\theta$.

\subsection{Supervision}
Learning supervisor parameters $\phi$ is important for supervising the student network.
Basically, the supervisor learns to maximize the likelihood,
e.g., $\log P(\mathbf{y} - \hat{\mathbf{y}}_s | \mathbf{x}_1, \mathbf{x}_2, \hat{\mathbf{y}}_s, \phi)$,
where the student output $\hat{\mathbf{y}}_s$ is inferred from 
an augmented input $\tilde{\mathbf{x}}_1, \tilde{\mathbf{x}}_2$.
First, it is natural to give the conditional knowledge from the labeled data $p_s$.
For this purpose, we minimize $\mathcal{L}_\text{TS}$ -- which stands for
supervised teacher loss -- to train the supervisor:
\begin{equation}
    \mathcal{L}_\text{TS}(\phi) = \mathbb{E}_{(\mathbf{x}_1, \mathbf{x}_2, \mathbf{y})\sim p_s}[\ell_\text{sup}(f_\phi(\mathbf{x}_1, \mathbf{x}_2, f_\theta(\tilde{\mathbf{x}}_1, \tilde{\mathbf{x}}_2)), \mathbf{y})].
\end{equation}
In addition to using the labeled data, we found that 
if a labeled dataset and an unlabeled dataset are distant, e.g., Things $\leftrightarrow$ KITTI,
using the unsupervised loss is especially effective.
The unsupervised teacher loss $\mathcal{L}_\text{TU}$ on unlabeled data $p_u$ is defined by:
\begin{equation}
    \mathcal{L}_\text{TU}(\phi) = \mathbb{E}_{(\mathbf{x}_1, \mathbf{x}_2)\sim p_u}[\ell_\text{unsup}(f_\phi(\mathbf{x}_1, \mathbf{x}_2, \hat{\mathbf{y}}_s), \mathbf{x}_1, \mathbf{x}_2)].
\end{equation}

In the pretraining stage,
we train the student model from scratch using
the supervised loss $\mathcal{L}_\text{sup}$, 
resulting in a pretrained weight $\theta$.
In the fine tuning stage, we initialize $\phi$ with $\theta$ and
jointly optimize $\theta$ and $\phi$ 
on labeled and unlabeled datasets.
Formally, we use stochastic gradient descent to minimize 
\begin{equation}
\mathcal{L}(\theta, \phi) = \mathcal{L}_\text{FS}(\theta) + 
\lambda_\text{TS}\mathcal{L}_\text{TS}(\phi) + \lambda_\text{TU}\mathcal{L}_\text{TU}(\phi),\label{eq:finalloss}
\end{equation}
with hyper-parameter loss weights $\lambda_\text{TS}$ and $\lambda_\text{TU}$.

\section{Experiments}

\subsection{Experimental setup}\label{sec:exprsetup}
\paragraph{Pretraining.} 
Our pretraining stage follows the original RAFT~\cite{teed2020raft}.
We pretrain our student network with FlyingChairs~\cite{dosovitskiy2015flownet} and FlyingThings3D~\cite{ilg2017flownet} with random cropping, scaling, color jittering, and block erasing.
The pretraining scheme includes 100k steps for FlyingChairs and additional 100k steps for FlyingThings3D, with the same learning rate schedule of RAFT, on which we base our network.
For the supervised loss (Eq.~\ref{eq:suploss}), we use $L_1$ loss.

\paragraph{Flow supervisor.} 
The flow supervisor is implemented with the GRU update module of
RAFT, which performs an iterative refinement process 
with the output flow of the previous step.
At the start of the semi-supervised training phase,
we initialize the supervisor model with the pretrained weights
of the GRU update module.
To match the student prediction from cropped inputs 
to the supervisor module with full resolution, i.e., privileged,
we pad the student predictions with zero.
In a training and testing phase, 
we use 12 iterations for both the student and supervisor GRUs.

\paragraph{Semi-supervised dataset.}
To compare semi-supervised learning
performance on Sintel~\cite{Butler:ECCV:2012} and KITTI~\cite{menze2015object}, we follow
the protocol~\cite{jonschkowski2020matters},
which utilizes the unlabeled portion, i.e., testing set,
of each dataset for training;
the difference is that we use a labeled dataset,
e.g., FlyingThings3D, into training.
Specifically, we use the unlabeled portion of Sintel for fine tuning on Sintel and unlabeled KITTI multiview dataset for fine tuning on KITTI; then the labeled splits are used for evaluation.

\paragraph{Optimization.}
We initialize our student and supervisor models with
the pretrained weights, then minimize 
the joint loss (Eq.~\ref{eq:finalloss}) for 100k steps.
We use $\alpha=1.0$, $\lambda_\text{TS} = 1.0$  and $\lambda_\text{TU} = 0$ by default  and $\lambda_\text{TU} = 0.01$ for the Things + KITTI setting, unless otherwise stated.
Detailed optimization settings and hyper-parameters are provided in the supplementary material and the code.

\begin{figure}[t]
\begin{subfigure}[b]{0.41\linewidth}
 \centering
 \includegraphics[width=\linewidth]{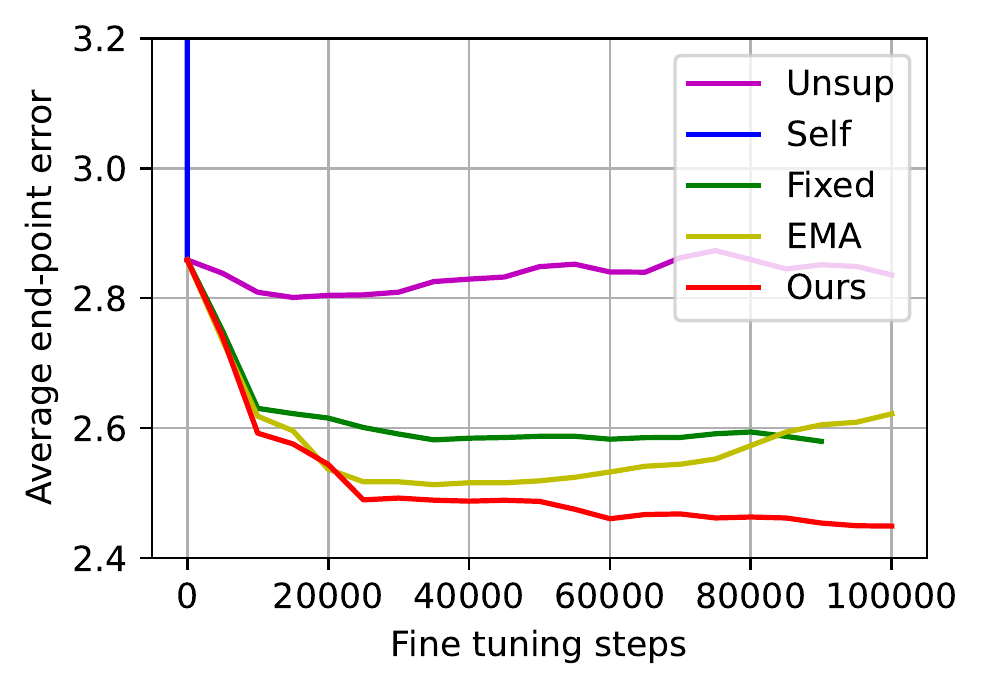}
 \caption{Sintel Final}
 \end{subfigure}
\begin{subfigure}[b]{0.39\linewidth}
 \centering
 \includegraphics[width=\linewidth]{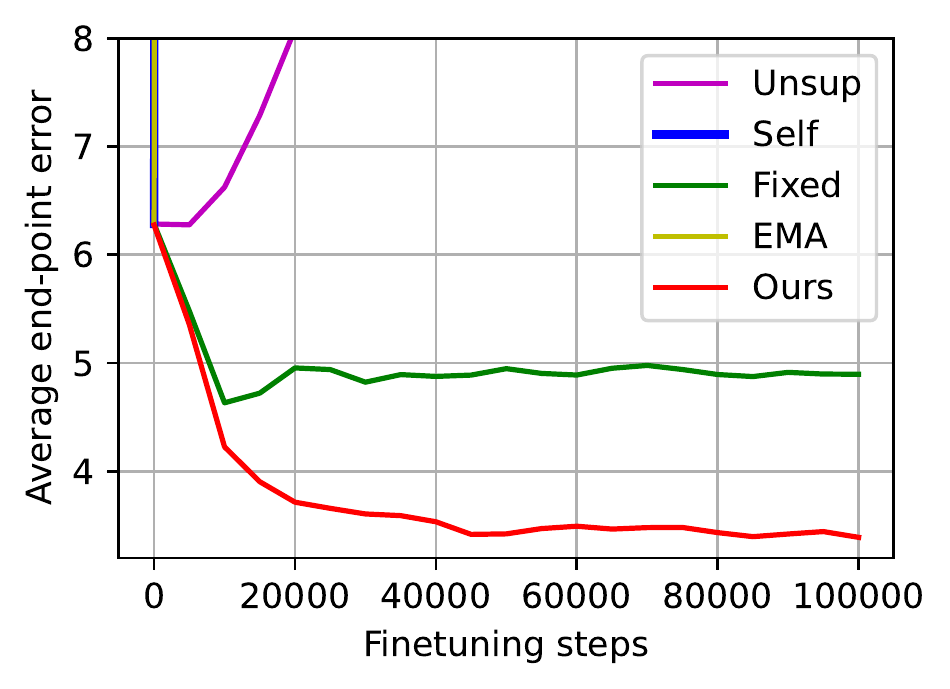}
 \caption{KITTI}
\end{subfigure}
\centering
\caption{\textbf{Plots comparing finetuning stage}.
We use FlyingThings3D as labeled data, and each target dataset as an unlabled data.
The EPEs are measured on unseen portion of data}
\label{fig:da_result}
\end{figure}

\subsection{Empirical Study}
\subsubsection{Comparison to Semi-Supervised Baselines.}

In Table~\ref{tab:da_result} and Fig.~\ref{fig:da_result}, 
we perform an empirical study, comparing ours with 
several baselines for semi-supervised learning. 
In the experiments, we use FlyingThings3D as the labeled data $p_s$ and
each target dataset as the unlabeled data $p_u$.

\begin{wraptable}[16]{rt}{0.5\linewidth}
\centering
\small
\caption{\textbf{Comparison to semi-supervised baselines.} 
Semi-supervised baselines and ours are compared,
as well as the supervised loss only (Sup) result.
We use widely-used metrics in optical flow: end-point error (EPE) and ratio of erroneous pixels (Fl)}\label{tab:da_result}
\resizebox{1.0\linewidth}{!}{%
\begin{tabular}{lcccc}
\hline
\multirow{2}{*}{Method} &  \multicolumn{2}{c}{Sintel} & \multicolumn{2}{c}{KITTI} \\
 & \multicolumn{1}{c}{Clean} & \multicolumn{1}{c}{Final} & \multicolumn{1}{c}{EPE} & \multicolumn{1}{c}{Fl-all (\%)}\\\hline\hline
Sup &  1.46 & 2.80 &  5.79 & 18.79\\\hline
Sup + Unsup  & 1.47  & 2.73 &  9.21 & 16.95 \\
Sup + Self &  \multicolumn{2}{c}{diverged} & \multicolumn{2}{c}{diverged} \\
Sup + EMA & 1.40 & 2.63 &  \multicolumn{2}{c}{diverged} \\
Sup + Fixed & 1.32 & 2.58 & 4.91 & 15.92 \\
\textbf{Sup + FS (Ours)} &  \textbf{1.30} & \textbf{2.46} & \textbf{3.35} & \textbf{11.12} \\\hline 
\end{tabular}
}
\end{wraptable}

\paragraph{Unsupervised loss.}
An unsupervised loss is designed to learn optical flow
without labels.
Exploring the unsupervised loss 
is a good start for the research
since we could expect a meaningful supervision signal
from unlabeled images.

In the second row of Table~\ref{tab:da_result},
we report results by the unsupervised loss.
We use the loss function in Eq.~\ref{eq:unsuploss},
which includes the census transform, full-image warping, smoothness prior, and occlusion handling as used in the prior work~\cite{stone2021smurf}.
Unfortunately, the unsupervised loss function does not give a
meaningful supervision signal when jointly used
with the supervised loss, resulting in a degenerated EPE on KITTI (5.79 $\rightarrow$ 9.21).
Interestingly, applying the identical loss to our supervisor, i.e., $\mathcal{L}_\text{TU}$, results in a better EPE on KITTI (5.79 $\rightarrow$ 3.35),
see also Table~\ref{tab:ablation}.

\paragraph{Self- and teacher-supervision.}
We summarize the results of the self-supervision and teacher-based methods in Table~\ref{tab:da_result} and Fig.~\ref{fig:da_result}.
The baseline models include the plain self-supervision (Self),
the fixed teacher (Fixed), and the EMA teacher (EMA).
In the plain self-supervision, we use the plain siamese networks (Fig.~\ref{subfig:rt_concept_a}).
The fixed teacher~\cite{zoph2020rethinking} and 
the EMA teacher~\cite{tarvainen2017mean}
are inspired by the existing literature. 
The self-supervision loss used in the experiments is defined by:
\begin{equation}
        \ell_\text{self} = \rho[f_\theta(\tilde{\mathbf{x}}_1, \tilde{\mathbf{x}}_2) - \texttt{stop\_grad}(f_t(\mathbf{x}_1, \mathbf{x}_2))],
\end{equation}
where $f_t$ is the teacher network of each baseline:
$t = \theta$ for plain self-supervision,
$t = \text{EMA}(\theta)$ for mean teacher, and
$t = \theta_\text{pretrained}$ for fixed teacher.
The plain self-supervision (Self) quickly diverges 
during the early fine tuning step for both Sintel and KITTI.
In the EMA teacher (EMA) results, more stable convergence is observed for Sintel, not KITTI.
We speculate that this unstable convergence is caused by the domain gap between the unlabeled data and the labeled data;
there exists a wider domain difference between
FlyingThings3D ($p_s$)
and KITTI ($p_u$)
than the difference between 
FlyingThings3D ($p_s$)
and Sintel ($p_u$),
since KITTI is a real-world dataset,
while Things and Sintel are both three-dimensional rendered datasets.
When the fixed teacher (Fixed) is used, we can observe more stable learning.
Our method (FS), enabling the supervisor learning along with the student, shows 
superior semi-supervised performance on both datasets
than the EMA and the fixed teacher.
We further analyze our method in the following sections.

\begin{table}[t]
\caption{\textbf{Comparison to optical flow approaches.} We report a percentage of improvement over each baseline in the parentheses. We mark used datasets: FlyingChairs (C), FlyingThings (T), unlabeled KITTI (K), and AutoFlow (A)~\cite{sun2021autoflow}}
    \label{tab:my_label}
    \centering
    \begin{subtable}[h]{0.5\textwidth}
    \caption{\textbf{Comparison to SemiFlowGAN~\cite{lai2017semi}}
    }
    \label{tab:ganflow}
    \resizebox{\textwidth}{!}{%
    \begin{tabular}{clll}
    \hline
    \multirow{2}{*}{Data} &
    \multirow{2}{*}{Method} & \multicolumn{2}{c}{KITTI} \\
     &  & \multicolumn{1}{c}{EPE} & \multicolumn{1}{c}{Fl-all (\%)}\\\hline\hline
    C    & SFGAN~\cite{lai2017semi} & 17.19 & 40.82\\
    C+K    & SFGAN~\cite{lai2017semi} & 16.02 {\small(-6.8\%)} & 38.77 {\small(-5.0\%)} \\\hline
    C+T & RAFT & 5.79 & 18.79\\
    T+K & RAFT + FS & \textbf{3.35 {\small(-42.1\%)}} & \textbf{11.12 {\small(-40.8\%)}} \\\hline 
    \\
    \end{tabular}
    }
    \end{subtable}\hfill
    \begin{subtable}[h]{0.48\textwidth}
    \caption{\textbf{Comparison to AutoFlow~\cite{sun2021autoflow} } 
    }
    \label{tab:autoflow}
    \small
    \resizebox{\textwidth}{!}{%
    \begin{tabular}{lcccc}
    \hline
    \multirow{2}{*}{Method} &  \multicolumn{2}{c}{Sintel} & \multicolumn{2}{c}{KITTI} \\
     & \multicolumn{1}{c}{Clean} & \multicolumn{1}{c}{Final} & \multicolumn{1}{c}{EPE} & \multicolumn{1}{c}{Fl-all (\%)}\\\hline\hline
    RAFT$^\dagger$ (C+T)~\cite{sun2021autoflow} &  1.68 & 2.80 &  5.92 & -\\
    RAFT (C+T) &  1.46 & 2.80 &  5.79 & 18.79\\\hline
    RAFT$^\dagger$ (A)~\cite{sun2021autoflow} &  1.95 & 2.57 &  4.23 & -\\
    \textbf{RAFT + FS (Ours)} &  \textbf{1.30} & \textbf{2.46} & \textbf{3.35} & \textbf{11.12} \\\hline 
    \multicolumn{5}{l}{$\dagger$ model implemented by \cite{sun2021autoflow}}\\
    \end{tabular}
    }
    \end{subtable}
    
\end{table}

\subsubsection{Comparison to SemiFlowGAN~\cite{lai2017semi}.}
We compare our method with the existing semi-supervised optical flow method~\cite{lai2017semi} in Table~\ref{tab:ganflow}.
SemiFlowGAN uses a domain adaptation-like approach by
matching distributions of the error maps from each domain.
Compared to SemiFlowGAN, our approach gives more direct supervision
to optical flow predictions from the supervisor model.
Here, we evaluate
how much a supervised-only baseline, i.e., trained w/o KITTI, is able to be 
improved by exploiting the unlabeled target dataset, i.e. traind w/ KITTI.
We can clearly observe much larger performance improvement (-6.8\% vs. -42.1\%)
when our method is used.

\subsubsection{Comparison to AutoFlow~\cite{sun2021autoflow}.}
In Table~\ref{tab:autoflow}, we compare ours with AutoFlow, where
our method shows better EPEs on both Sintel and KITTI.
AutoFlow is devised to train a better neural network on target 
datasets by learning to generate data,
as opposed to ours using semi-supervised learning.
For instance, the AutoFlow dataset, whose generator is optimized on Sintel dataset
shows superiority over a traditional synthetic dataset, e.g., FlyingThings, in generalization on unseen target domains.
On the other hand, our strategy is to utilize a synthetic dataset and an unlabeled target domain dataset
by the supervision of the flow supervisor; it boosts performance on the target domain without labels.

\subsubsection{Ablation Study.}\label{sec:ablation_study}

\begin{wraptable}[16]{r}{0.5\linewidth}
\caption{\textbf{Ablation experiments.} We underline the final settings}\label{tab:ablation}
\small
\centering
\resizebox{1.0\linewidth}{!}{%
\begin{tabular}{lccccc}
\hline
\multirow{2}{*}{Experiment} &
\multirow{2}{*}{} &  \multicolumn{2}{c}{Sintel} & \multicolumn{2}{c}{KITTI} \\
& & \multicolumn{1}{c}{Clean} & \multicolumn{1}{c}{Final} & \multicolumn{1}{c}{EPE} & \multicolumn{1}{c}{Fl-all}\\\hline\hline
Parameter & \underline{on} &  \textbf{1.30} & \textbf{2.46} & \textbf{3.35} & \textbf{11.12} \\
separation & off & \multicolumn{2}{c}{diverged} & \multicolumn{2}{c}{diverged} \\\hline
Passing & \underline{full} &  \textbf{1.30} & \textbf{2.46} & \textbf{3.35} & \textbf{11.12} \\
student & w/o res & 1.34 & 2.50 & 6.16 & 12.34 \\
output & off & \textbf{1.30} & \textbf{2.46} & 6.68 & 13.36 \\\hline
Shared  & \underline{on}  & \textbf{1.30} & \textbf{2.46} & \textbf{3.35} & \textbf{11.12} \\
encoder & off & 1.33 & 2.58 & 3.80 & 12.03\\\hline
Teacher  & \underline{TS} & \underline{\textbf{1.30}} & \underline{\textbf{2.46}} & 4.69 & 14.48\\
loss type & \underline{+TU} & 1.55 & 2.80 & \underline{\textbf{3.35}} & \underline{\textbf{11.12}} \\\hline
Teacher  & \underline{clean} &  \textbf{1.30} & \textbf{2.46} & \textbf{3.35} & \textbf{11.12} \\
input & aug & 1.33 & 2.56 & 4.17 & 11.55 \\\hline
\end{tabular}}
\end{wraptable}

In Table~\ref{tab:ablation}, we compare ours with several alternatives.

\paragraph{Parameter separation.}
Self-supervision with the shared network (see Self in Fig.~\ref{fig:da_result}) 
is unsuitable for semi-supervised optical flow learning.
On the other hand, having the separate parameters for the supervisor network
effectively prevents the divergence. 
This separation can be viewed as the predictor strategy in the self-supervised learning context~\cite{chen2021exploring}.
We also analyze the separate network as 
a meta-learner, which learns knowledge specifically for 
supervision. We have shortly discussed this aspect in Sec.~\ref{subsec:residual_teacher}.

\paragraph{Passing student outputs ($\hat{\mathbf{y}}_s, \mathbf{h}_s$).}
Our architecture passes the student output flow and internal state to the supervisor network to enable the supervisor to learn the residual function, as described in Eq.~\ref{eq:residual}.
In row 5 in Table~\ref{tab:ablation}, 
we show the results when we do not pass student outputs ($\hat{\mathbf{y}}_s, \mathbf{h}_s$) to the teacher.
Interestingly, the Sintel results remain the same even without passing student outputs.
while the KITTI results benefit from passing student outputs.
We ablate the residual connection from the network, 
while still passing student outputs to the supervisor (w/o res).
In this case (w/o res), we can observe 
the better KITTI results over the no connection case (off),
but worse than our full design.
These results indicate that conditioning 
supervisor with the student helps
find residual function  $\Delta f_\phi(\hat{\mathbf{y}}_s) \approx \mathbf{y} -\hat{\mathbf{y}}_s$,
especially on distant domains, i.e., Things $\leftrightarrow$ KITTI.
Overall, passing with residual connection (full) performs
consistently better.

\paragraph{Shared encoder.}
Shared encoder design results in a better flow accuracy,
as shown in Table~\ref{tab:ablation}.
Our method
uses the shared encoder design (Fig.~\ref{subfig:rt_concept_b}).
Separating the encoder results in worse EPEs 
for Sintel and KITTI.
Not only the accuracy, but it also results in a less efficient training pipeline due to the increased number of parameters by the encoder.

\paragraph{Teacher loss type.}
We propose to train the supervisor network 
with labeled and unlabeled data.
Supervised teacher (TS) loss utilizes the labeled dataset
for supervisor learning. 
In both datasets, TS improves the baseline performance.
We additionally apply the unsupervised teacher loss (+TU), which results in better EPE in KITTI but
is ineffective for Sintel.
This is related to the results of pure unsupervised learning~\cite{stone2021smurf},
where the loss is shown to be more effective on KITTI than on Sintel.

\begin{wraptable}[7]{r}{0.5\linewidth}
\caption{\textbf{KITTI results of models trained on VKITTI~\cite{cabon2020vkitti2}}}
\label{tab:vkitti}
\small
\centering
\resizebox{0.95\linewidth}{!}{%
\begin{tabular}{cccc}
\hline
 & \multicolumn{2}{c}{Supervised} & Semi-supervised \\
Metric & RAFT & SepFlow~\cite{zhang2021separable} & RAFT+FS (ours) \\
\hline
EPE & 3.64 & 2.60 & \textbf{2.39}\\
Fl-all (\%)  & 8.78 & 7.74 & \textbf{7.63} \\\hline
\end{tabular}}

\end{wraptable}

\subsubsection{Virtual KITTI.}
With our method, we can utilize a virtual driving dataset, VKITTI~\cite{cabon2020vkitti2}, for training a better model
for the real KITTI dataset.
In Table~\ref{tab:vkitti},
we show results obtained by VKITTI.
In the supervised case, 
our base network (RAFT) shows 3.64 EPE on KITTI,
and SeperableFlow~\cite{zhang2021separable} shows 2.60 EPE.
In the semi-supervised case, our method leverages the unlabeled KITTI multiview set
additional to VKITTI, and we train the RAFT network for 50k steps; 
the resulting network performs better on KITTI.

\subsubsection{Supervisor vs. student.}
\begin{wrapfigure}[14]{r}{0.58\linewidth}
\centering
\begin{subfigure}[b]{0.49\linewidth}
    \centering
    \includegraphics[width=\linewidth]{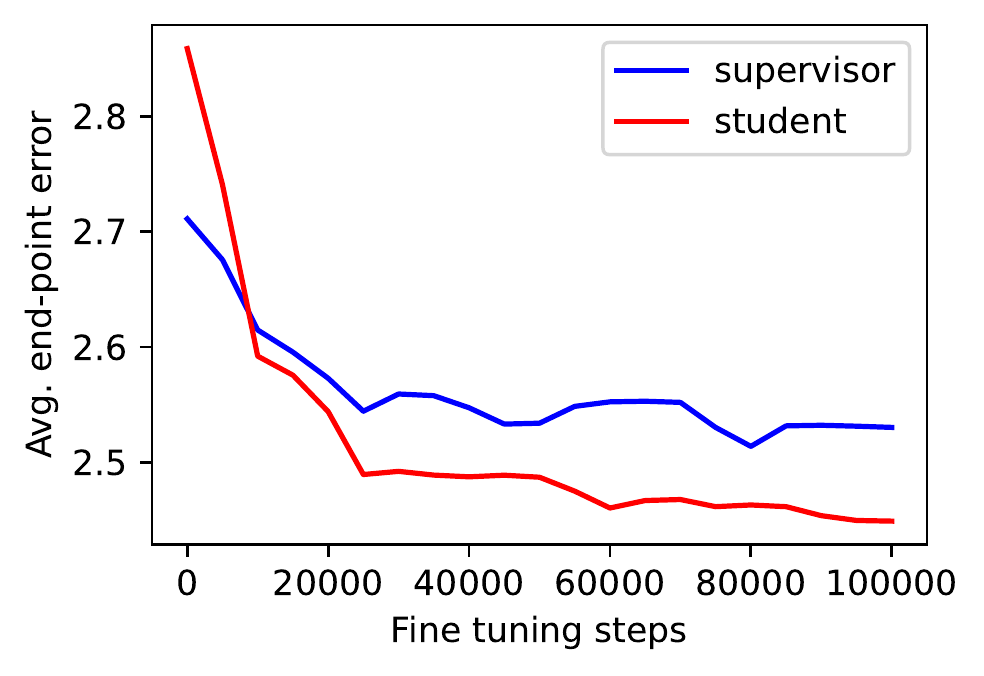}
    \caption{Sintel}
\end{subfigure}
\begin{subfigure}[b]{0.49\linewidth}
    \centering
    \includegraphics[width=\linewidth]{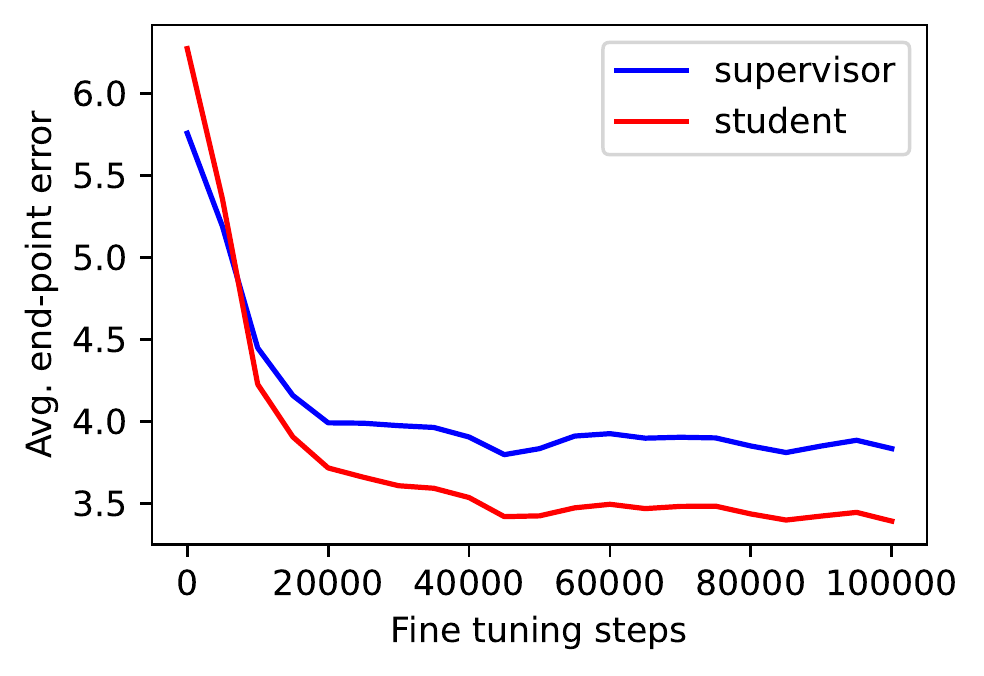}
    \caption{KITTI}
\end{subfigure}
\caption{\textbf{Supervisor vs. student EPEs} during semi-supervised fine tuning. EPEs measured on unseen portion of data}
\label{fig:tvs}
\end{wrapfigure}
In Fig.~\ref{fig:tvs}, we show the performance of the student and the
supervisor networks by training steps.
We use clean inputs for both the student and the supervisor networks for evaluation.
Interestingly, we observe that the student
is better than the supervisor during fine tuning.
In addition, we could observe EPEs of the student and the supervisor are correlated, and they are both improved by our semi-supervised training.
In knowledge distillation, we can 
observe similar behavior~\cite{stanton2021does},
where a student model shows better validation accuracy than a teacher model. 

\begin{figure*}[p]
\centering
\begin{subfigure}[b]{0.21\linewidth}
 \centering
 \includegraphics[trim={0mm 140mm 213.25mm 25.9mm},clip,width=\linewidth]{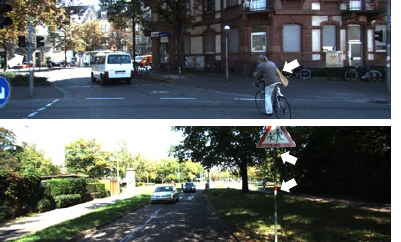}
 \caption{\small Input}
 \end{subfigure}
\begin{subfigure}[b]{0.21\linewidth}
 \centering
 \includegraphics[trim={40.75mm 140mm 172.5mm 25.9mm},clip,width=\linewidth]{figure/qual_combined.pdf}
 \caption{\small Sup on KITTI}
\end{subfigure}
\begin{subfigure}[b]{0.21\linewidth}
 \includegraphics[trim={81.5mm 140mm 131.75mm 25.9mm},clip,width=\linewidth]{figure/qual_combined.pdf}
 \caption{\small Ours}
 \end{subfigure}
\begin{subfigure}[b]{0.21\linewidth}
 \centering
  \includegraphics[trim={122.25mm 140mm 91mm 25.9mm},clip,width=\linewidth]{figure/qual_combined.pdf}
 \caption{\small Ground truth}
\end{subfigure}
\centering
\caption{\textbf{Qualitative results on KITTI.} We visualize optical flow predicted by \textbf{(b)} supervised on target dataset and \textbf{(c)} semi-supervised w/o target label.
Note that sparse ground-truth \textbf{(d)} is not sufficient to make a clear boundary of objects (marked with arrows), while our method shows better results
}
\label{fig:qual_test}
\end{figure*}

\begin{figure*}[p]
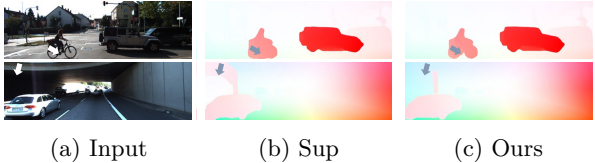

\centering
\begin{subfigure}[b]{0.21\linewidth}
    \centering
    \includegraphics[trim={0mm 165.9mm 213.25mm 0mm},clip,width=\linewidth]{figure/qual_combined.pdf}
    \caption{Input}
\end{subfigure}
\begin{subfigure}[b]{0.21\linewidth}
    \centering
    \includegraphics[trim={40.75mm 165.9mm 172.5mm 0mm},clip,width=\linewidth]{figure/qual_combined.pdf}
    \caption{Sup}
\end{subfigure}
\begin{subfigure}[b]{0.21\linewidth}
    \includegraphics[trim={81.5mm 165.9mm 131.75mm 0mm},clip,width=\linewidth]{figure/qual_combined.pdf}
    \caption{Ours}
\end{subfigure}

\caption{\textbf{Qualitative results on KITTI testing samples.} We compare the supervised model (Sup) with the semi-supervised model (Ours). Both models 
exploit KITTI labels; ours utilizes additional unlabeled KITTI for fine tuning}
\label{fig:qual_kitti}
\end{figure*}

\begin{figure*}[p]
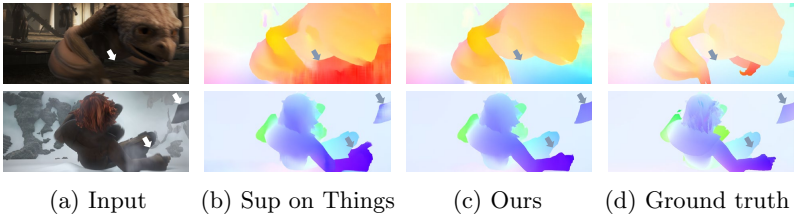

\centering
\begin{subfigure}[b]{0.21\linewidth}
 \centering
 \includegraphics[trim={0mm 102.3mm 213.25mm 53mm},clip,width=\linewidth]{figure/qual_combined.pdf}
 \caption{Input}
 \end{subfigure}
\begin{subfigure}[b]{0.21\linewidth}
 \centering
 \includegraphics[trim={40.75mm 102.3mm 172.5mm 53mm},clip,width=\linewidth]{figure/qual_combined.pdf}
 \caption{Sup on Things}
\end{subfigure}
\begin{subfigure}[b]{0.21\linewidth}
 \includegraphics[trim={81.5mm 102.3mm 131.75mm 53mm},clip,width=\linewidth]{figure/qual_combined.pdf}
 \caption{Ours}
 \end{subfigure}
\begin{subfigure}[b]{0.21\linewidth}
 \centering
  \includegraphics[trim={122.25mm 102.3mm 91mm 53mm},clip,width=\linewidth]{figure/qual_combined.pdf}
 \caption{Ground truth}
\end{subfigure}
\centering
\caption{\textbf{Qualitative results on Sintel.} We visualize optical flow predicted by \textbf{(b)} supervised on FlyingThings3D and \textbf{(c)} semi-supervised without target label. Though ours trained without target labels, it successfully adapts the pretrained model to the target domain. Improved areas are marked by arrows}
\label{fig:qual}
\end{figure*}

\begin{figure}[p]
\centering
\begin{subfigure}[b]{0.18\linewidth}
 \centering
 \includegraphics[trim={0mm 162.6mm 204.5mm 0mm},clip,width=\linewidth]{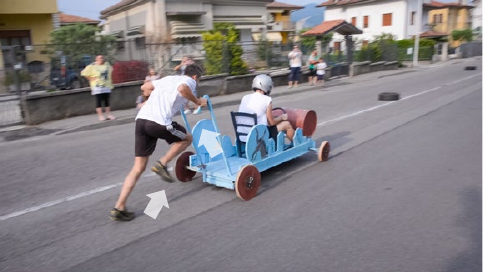}
 \includegraphics[trim={0mm 106.7mm 204.5mm 56.3mm},clip,width=\linewidth]{figure/qual/davis_low}
 \caption{\tiny Input}
 \end{subfigure}
\begin{subfigure}[b]{0.18\linewidth}
 \centering
 \includegraphics[trim={50mm 162.6mm 154.5mm 0mm},clip,width=\linewidth]{figure/qual/davis_low}
 \includegraphics[trim={50mm 106.7mm 154.5mm 56.3mm},clip,width=\linewidth]{figure/qual/davis_low}
 \caption{\tiny CT (Sup)}
\end{subfigure}
\begin{subfigure}[b]{0.18\linewidth}
 \includegraphics[trim={100mm 162.6mm 104.5mm 0mm},clip,width=\linewidth]{figure/qual/davis_low}
 \includegraphics[trim={100mm 106.7mm 104.5mm 56.3mm},clip,width=\linewidth]{figure/qual/davis_low}
 \caption{\tiny CT (\textbf{Ours})}
 \end{subfigure}
\begin{subfigure}[b]{0.18\linewidth}
 \centering
  \includegraphics[trim={150mm 162.6mm 54.5mm 0mm},clip,width=\linewidth]{figure/qual/davis_low}
  \includegraphics[trim={150mm 106.7mm 54.5mm 56.3mm},clip,width=\linewidth]{figure/qual/davis_low}
 \caption{\tiny CTSKH (Sup)}
\end{subfigure}
\begin{subfigure}[b]{0.18\linewidth}
 \centering
  \includegraphics[trim={200mm 162.6mm 4.5mm 0mm},clip,width=\linewidth]{figure/qual/davis_low}
  \includegraphics[trim={200mm 106.7mm 4.5mm 56.3mm},clip,width=\linewidth]{figure/qual/davis_low}
 \caption{\tiny CTSKH (\textbf{Ours})}
\end{subfigure}
\centering
\caption{\textbf{Qualitative results on DAVIS dataset~\cite{Perazzi2016}.} We fine tune each pretrained network (Sup) on Davis dataset by our semi-supervised method (Ours). Improved regions are marked with arrows. These optical flows are inferred on unseen portion of the dataset}
\label{fig:qual_davis}
\end{figure}

\subsection{Qualitative results}
We provide qualitative results on KITTI (Fig.~\ref{fig:qual_test}, \ref{fig:qual_kitti}) and Sintel (Fig.~\ref{fig:qual}).
Since the KITTI dataset includes sparse ground truth flows,
training by the ground truth supervision
often results in incorrect results, especially in
boundaries and deformable objects (e.g., person on a bike).
Thus, in this case, we can expect better flows by exploiting semi-supervised methods.
The Sintel Final dataset includes challenging blur, fog, and lighting conditions as shown in the examples.
Interestingly, the semi-supervision without a target dataset label helps improve the challenging regions, when our method is used.

In Fig.~\ref{fig:qual_davis}, we show results on DAVIS dataset~\cite{Perazzi2016} by the pretrained models
and our fine tuned models; the dataset does not contain any optical flow ground truth.
From the training set with 3,455 frames, we utilize 90\% of frames for training and
the rest for qualitative evaluation.
Even though our fine tuning does not utilize ground truth,
we can observe a clear positive effect in several challenging regions.
Especially, our method improves blurry regions (e.g., moving hand of the dancer) and object boundaries.

\subsection{Comparison to State-of-the-arts}\begin{table*}[t]
\centering \small
\caption{\textbf{Comparison to state-of-the-arts.} 
Data usage is abbreviated to FlyingChairs (C), FlyingThings3D (T), Sintel (S), KITTI (K), HD1K~\cite{kondermann2016hci} (H), Sintel unlabeled ($\text{S}^u$), KITTI unlabeled ($\text{K}^u$), and Spring ($\text{Spg}^u$).
For labeled datasets, we follow the training scheme detailed in each paper. RAFT\textsuperscript{tf} is our implementation in TensorFlow}\label{tab:sota_comparison}
\resizebox{0.95\textwidth}{!}{%
\begin{tabular}{lllccccccc}
\hline
Labeled                    & Unlabeled                                 &                    & \multicolumn{2}{c}{Sintel (train)} & \multicolumn{2}{c}{KITTI (train)} & \multicolumn{2}{c}{Sintel (test)} & KITTI (test) \\
data & data & Method & Clean & Final & EPE & Fl-all (\%) & Clean & Final & Fl-all (\%)\\\hline\hline
\multirow{5}{*}{C+T} & \multirow{4}{*}{-} & RAFT~\cite{teed2020raft} & 1.43 & 2.71 & 5.05 & 17.4 & - & - & - \\
& & RAFT\textsuperscript{tf} & 1.46            & 2.80            & 5.79            & 18.8              & -               & -              & -               \\
& & GMA~\cite{Jiang_2021_ICCV}                & \textbf{1.30}            & 2.74            & 4.69            & 17.1         & -               & -              & -               \\
& & SeparableFlow~\cite{zhang2021separable}      & \textbf{1.30}            & \underline{2.59}            & \underline{4.60}            & \underline{15.9}               & -               & -              & -               \\\cline{2-10}
& $\text{S}^u$/$\text{K}^u$  & RAFT\textsuperscript{tf}+FS (Ours) & \textbf{1.30}& \textbf{2.46} & \textbf{3.35} & \textbf{11.12}  & -  & -  & -  \\\hline
\multirow{5}{*}{C+T+S+K+H} & \multirow{3}{*}{-} & RAFT~\cite{teed2020raft} & (0.77) & (1.27) & (0.63) & (1.5) & 1.61 & 2.86 & 5.10 \\
&   & GMA~\cite{Jiang_2021_ICCV}       & (0.62)          & (1.06)          & (0.57)          & (1.2)              & \textbf{1.39}            & \underline{2.47}           & 5.15            \\
& & SeparableFlow~\cite{zhang2021separable}      & (0.69)          & (1.10)          & (0.69)          & (1.60)   & 1.50    & 2.67           & \textbf{4.64}\\\cline{2-10}
& \multirow{2}{*}{$\text{Spg}^u$/$\text{K}^u$} & RAFT+FS (Ours) & (0.75) & (1.29) &  (0.69)  & (1.75) &  1.65  & 2.79 & \underline{4.85}\\
& & GMA+FS (Ours)      &   (0.63) & (1.05)  & (0.61) & (1.47) & \underline{1.43} & \textbf{2.44} & 4.95 \\\hline       
\end{tabular}}

\end{table*}

\paragraph{Experimental settings.}
In this experiment we compare our method 
to the existing supervised methods.
The biggest challenge in supervised optical flow is that
we do not have ample 
ground truth flows for our target domains, e.g., 200 pairs labeled in KITTI.
Thanks to our semi-supervised method, 
we can train with related videos 
for better performance on the target datasets.
For GMA, we use $\alpha=0.25$ (Sintel) and
$\alpha=0.05$ (KITTI).

\paragraph{Dataset configuration.}
Results evaluated on the training sets of each dataset
are experimented with the setting described in Sec.~\ref{sec:exprsetup}.
On the other hand,
the results on the test splits of the two benchmarks
are obtained by justifiable external datasets
for semi-supervised training.
Though our method is able to utilize the unlabeled portion of Sintel, we 
bring another external set to avoid the relation to the testing samples, as follows.
To assist Sintel performance,
we use Spring (abbr. to $\text{Spg}^u$)~\cite{spring-blender}, which is the `open animated movie' by Blender Institute,
similar to Sintel~\cite{SintelMovie}.
In Spring, we use the whole frames (frame no. 1-11,138) without any modification.
Additionally, we use Sintel (train) with interval two as unlabeled.
For KITTI, we additionally use KITTI multiview dataset, which does not contain ground truth optical flows;
we use the training split of the dataset to avoid
duplicated scenes with testing samples.

\paragraph{Results.}
In Table~\ref{tab:sota_comparison}, 
we report the C+T result, which is a commonly used protocol to
evaluate the generality of models.
Since our method is designed to use unlabeled samples,
our method exploits each target dataset 
-- which is not overlapped with each evaluation sample -- without ground truth.
The results indicate we could achieve better accuracy
than the supervised-only approaches.
Interestingly, our approach performs better than the advanced architectures, i.e., GMA~\cite{Jiang_2021_ICCV} and SeperableFlow~\cite{zhang2021separable}, in the C+T category.

In `C+T+S+K+H', we report the test results with the external datasets.
In the KITTI test result where we have access to 200 labeled samples, using additional samples results in better Fl-all (5.10 $\rightarrow$ 4.85, 5.15 $\rightarrow$ 4.95).
For Sintel, we test on two base networks: RAFT and GMA.
In two networks, our method makes improvements
on Sintel Final (test).
However, our method does not improve accuracy on Sintel Clean set.
That is because we use the external Spring dataset,
which includes various challenging effects, e.g., blur, as Sintel Final,
the student model becomes robust to those effects rather than
clean videos.
Note that, for Sintel test,  
our improvement is made by the video of a different domain which is encoded by a lossy compression, like most videos on the web.

\subsection{Limitations}

A limitation of our approach is that it depends on a supervised baseline,
which is sometimes less preferable than an unsupervised approach.
A handful of unsupervised optical flow researches 
have achieved amazing performance improvement.
Especially, on KITTI (w/o target label), 
unsupervised methods have performed better than supervised methods,
and the recent performance gap has been widened (EPE: 2.0 vs 5.0).
Unfortunately, we observe that our method is not compatible with unsupervised baselines;
the semi-supervised fine tuning of SMURF~\cite{stone2021smurf} with our method (T+K$^u$)
results in a worse EPE (2.0 $\rightarrow$ 2.5) on KITTI.
Thus, one of the future work lies upon developing a fine-tuning method 
to improve the unsupervised baselines.

Nonetheless, our work contributes to the research by ameliorating
the limitation of a supervised method, suffering from worse generalization.
A supervised method is often preferable than an unsupervised one 
for higher performance even without target labels;
we show ours can be used in such cases.
For example, on Sintel Final, ours shows better EPE (2.46) than the supervised one~\cite{teed2020raft} (2.71) and the unsupervised method~\cite{stone2021smurf} (2.80).

\section{Conclusion}\label{sec:conclusion}

We have presented 
a self-supervision strategy for semi-supervised optical flow,
which is simple, yet effective.
Our flow supervisor module supervises a student model, which is effective in the semi-supervised optical flow setting where we have no or few samples in a target domain.
Our method outperforms various self-supervised baselines, shown by the empirical study.
In addition, we show that our semi-supervised method can improve the state-of-the-art supervised models by exploiting additional unlabeled datasets.

\subsubsection{Acknowledgment.}
{\small This work was supported by the NRF (2019R1A2C3002833) and IITP (IITP-2015-0-00199, Proximity computing and its applications to autonomous vehicle, image search, and 3D printing) grants funded by the Korea government (MSIT). We thank the authors of RAFT, GMA, and SMURF for their public codes and dataset providers of Chairs, Things, Sintel, KITTI, VKITTI, and Spring for making the datasets available.
Sung-Eui Yoon is a corresponding author.}

\bibliographystyle{splncs04}
\bibliography{mybib}

\clearpage
\appendix

\section{Additional Results}
\subsection{Refinement Steps and Faster Inference}
\begin{figure}[h]
\centering
\begin{subfigure}[b]{0.32\linewidth}
 \centering
 \includegraphics[width=\linewidth]{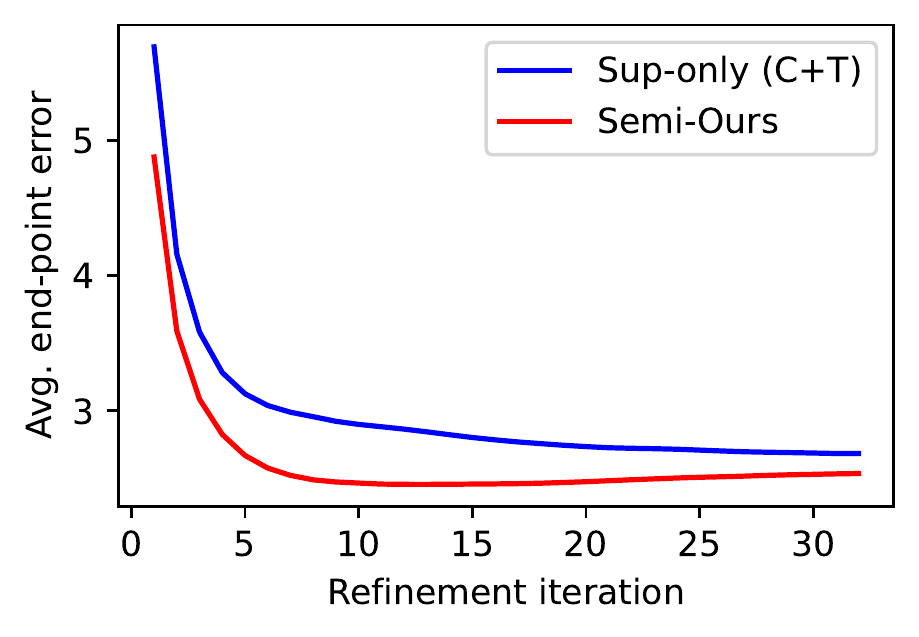}
 \caption{Sintel Final EPE}\label{subfig:iter_a}
 \end{subfigure}\hfill
\begin{subfigure}[b]{0.34\linewidth}
 \centering
 \includegraphics[width=\linewidth]{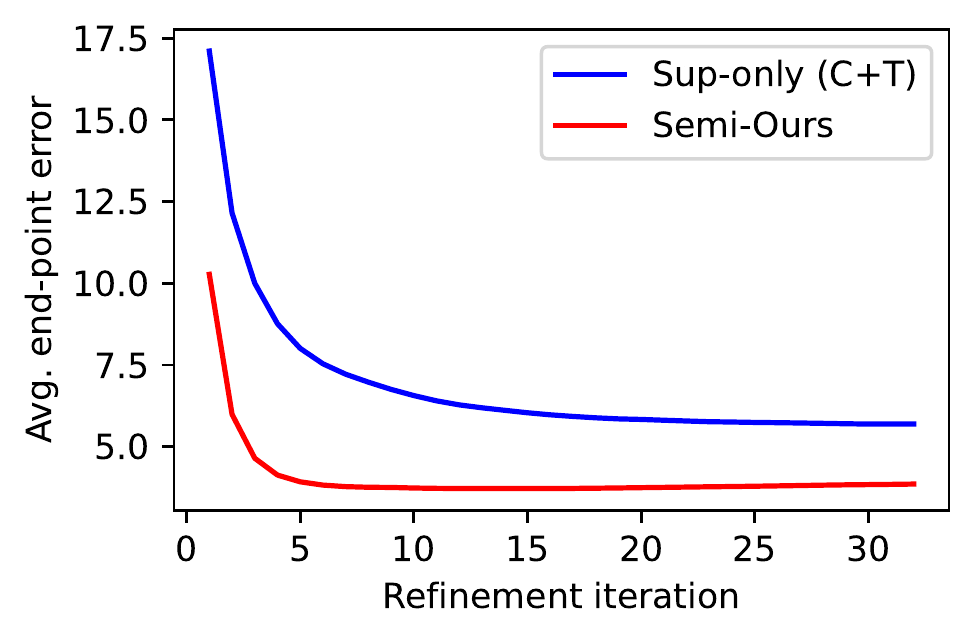}
 \caption{KITTI EPE}\label{subfig:iter_b}
 \end{subfigure}\hfill
\begin{subfigure}[b]{0.33\linewidth}
 \centering
 \includegraphics[width=\linewidth]{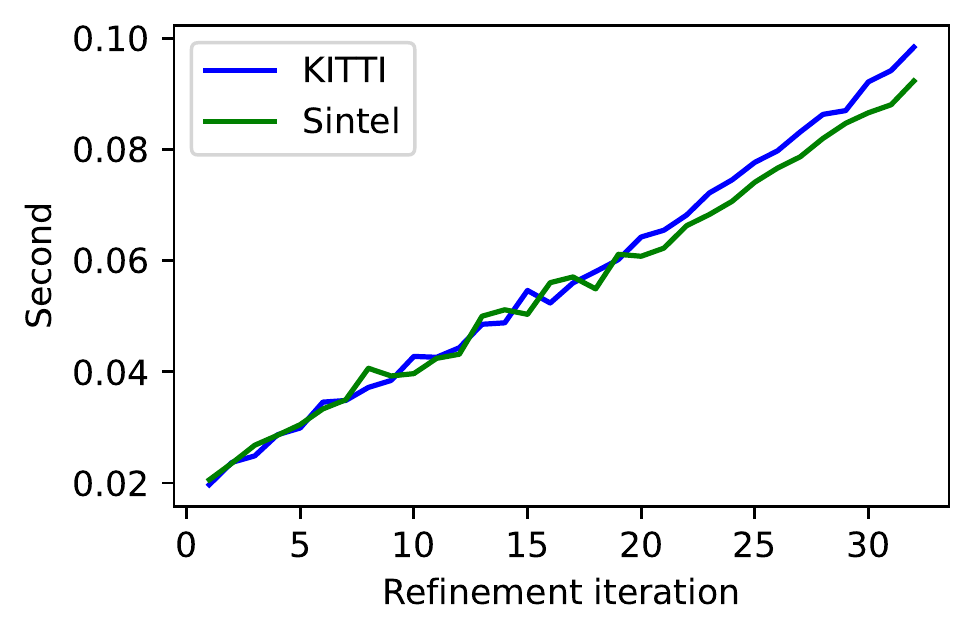}
 \caption{Time vs. Iteration}\label{subfig:iter_c}
 \end{subfigure}\hfill 
\centering
\caption{\textbf{Results w.r.t. the refinement iteration}. \textbf{(a-b)} shows validation EPEs of the baseline (Sup-only) and ours (Semi-Ours). \textbf{(c)} shows the inference time per frame with different resolutions: KITTI ($1242\times 375$) and Sintel ($1024\times 436$).
For the comparison of time, we use a single RTX 3090 GPU (24GB VRAM) with our TensorFlow implementation of RAFT. }
\label{fig:iter}
\end{figure}
In Fig.~\ref{fig:iter}, we show EPEs with respect to refinement steps.
Since we base our network on RAFT~\cite{teed2020raft}, setting an appropriate 
refinement step is crucial for the performance of estimation. 
By the experiments, we observe that training with our self-supervision method
results in a faster convergence.
In Fig.~\ref{subfig:iter_a}-\ref{subfig:iter_b},
we show that our semi-supervised learning method has a faster convergence iteration
than the baseline (Sup-only) model.
That is the reason we choose shorter refinement steps, i.e., 12,
than the original paper, i.e., 32. 
Thanks to the fewer refinement steps, the inference time is reduced by about 50\% (see Fig.~\ref{subfig:iter_c}) on a target dataset,
when we set the iteration to 12 instead of 32.

\subsection{Stopping Gradients}
Our design choice for \texttt{stop\_grad} is to make the supervisor module an isolated module,
which makes gradient computation step for the flow supervisor more compact and efficient.
Furthermore, stop-gradient is a widely adopted technique in self-supervised learning to prevent a degenerated solution~\cite{chen2021exploring}. 
We discovered that disabling  \textnormal{\texttt{stop\_grad}}
for each module shows worse results on Sintel Final:
\begin{table}[h]
\small
\vspace{-2.5mm}
    \centering
    \begin{tabular}{|c|c|c|c|}
    \hline
        Teacher encoder & $\mathbf{h}_s$ & $\hat{\mathbf{y}}_{FS}$ & \textbf{Ours} (all enabled) \\\hline
        2.58 & 2.52 & diverged & \textbf{2.46} \\\hline
    \end{tabular}
\end{table}\vspace{-4.5mm}

\noindent Note that \texttt{stop\_grad} for $\hat{\mathbf{y}}_s$
belongs to original RAFT.

\subsection{More Qualitative Results on Sintel and KITTI}
\begin{figure}[t]
\centering
\small
\begin{subfigure}[b]{0.23\linewidth}
 \centering
 \includegraphics[trim={0mm 84.9mm 204.1mm 0mm},clip,width=\linewidth]{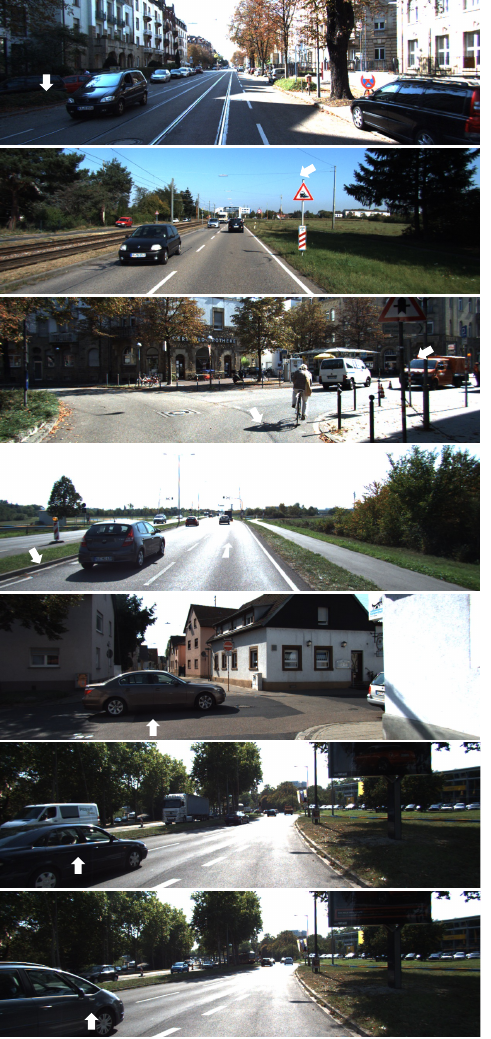}
 \caption{Input}
 \end{subfigure}\hfill
\begin{subfigure}[b]{0.23\linewidth}
 \centering
 \includegraphics[trim={50mm 84.9mm 154.1mm 0mm},clip,width=\linewidth]{figure/qual/vkittiours.pdf}
 \caption{Sup-only}
\end{subfigure}\hfill
\begin{subfigure}[b]{0.23\linewidth}
 \centering
 \includegraphics[trim={100mm 84.9mm 104.1mm 0mm},clip,width=\linewidth]{figure/qual/vkittiours.pdf}
 \caption{Semi-Ours}
\end{subfigure}\hfill
\begin{subfigure}[b]{0.23\linewidth}
 \centering
 \includegraphics[trim={150mm 84.9mm 54.1mm 0mm},clip,width=\linewidth]{figure/qual/vkittiours.pdf}
 \caption{Ground truth}
\end{subfigure}
\centering
\caption{\textbf{Qualitative results on KITTI.} We compare results of RAFT trained on VKITTI. \textbf{(b)} shows optical flows 
predicted by RAFT pretrained on VKITTI. \textbf{(c)} shows flows prediected by our semi-supervised method, which utilizes an additional KITTI
dataset without ground-truth. All results are obtained on unseen samples.}
\label{fig:vkittiours}
\end{figure}

\begin{figure}[t]
\centering
\begin{subfigure}[b]{0.23\linewidth}
 \centering
 \includegraphics[trim={0mm 62.8mm 204.1mm 0mm},clip,width=\linewidth]{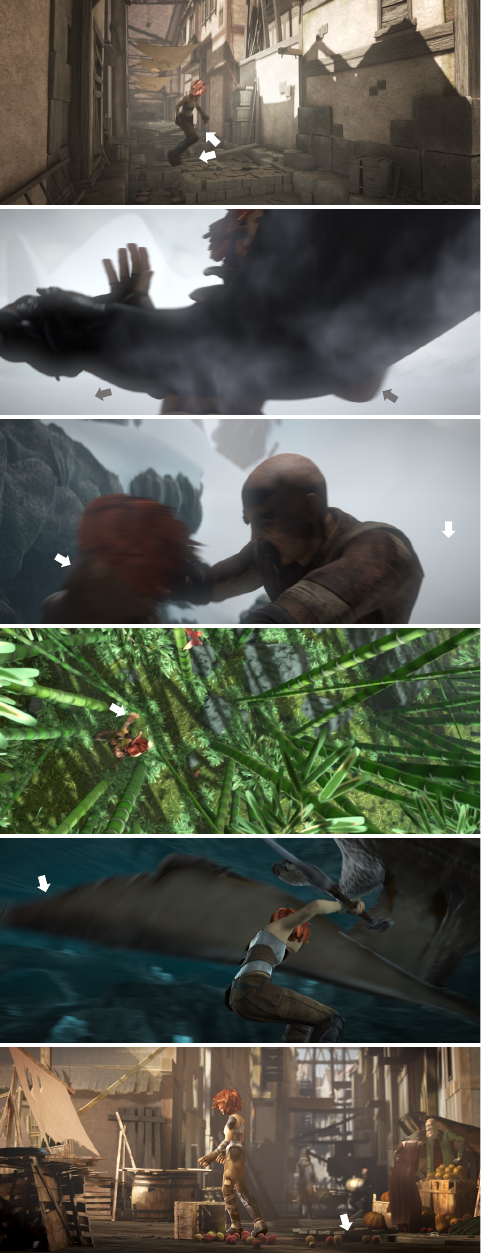}
 \caption{Input}
 \end{subfigure}\hfill
\begin{subfigure}[b]{0.23\linewidth}
 \centering
 \includegraphics[trim={50mm 62.8mm 154.1mm 0mm},clip,width=\linewidth]{figure/qual/suppsintel.pdf}
 \caption{C+T (Sup-only)}
\end{subfigure}\hfill
\begin{subfigure}[b]{0.23\linewidth}
 \centering
 \includegraphics[trim={100mm 62.8mm 104.1mm 0mm},clip,width=\linewidth]{figure/qual/suppsintel.pdf}
 \caption{{\tiny C+T} (Semi-Ours)}
\end{subfigure}\hfill
\begin{subfigure}[b]{0.23\linewidth}
 \centering
 \includegraphics[trim={150mm 62.8mm 54.1mm 0mm},clip,width=\linewidth]{figure/qual/suppsintel.pdf}
 \caption{Ground truth}
\end{subfigure}
\centering
\caption{\textbf{Qualitative results on Sintel Final.} We compare results of RAFT trained on C+T. \textbf{(b)} shows optical flows 
predicted by RAFT pretrained on FlyingChairs and FlyingThings. \textbf{(c)} shows flows prediected by our semi-supervised method, which utilizes an additional Sintel
dataset without ground-truth. All results are obtained on unseen samples.}\vspace{-10mm}
\label{fig:suppsintel}
\end{figure}

We show more qualitative comparisons
in Fig.~\ref{fig:vkittiours}-\ref{fig:suppsintel}.
Even though we do not exploit a ground truth of the target dataset,
our method clearly improves challenging regions.
In the KITTI examples, our method is especially effective on
objects near image frames and
shadows.
For Sintel -- which includes many motion blurs and lens effects --
the network becomes robust to 
those challenging effects, as
shown in the examples.

\section{Experimental Details}

\subsection{Architecture}
\begin{figure}[t]
\centering
\begin{subfigure}[b]{0.9\linewidth}
 \centering
 \includegraphics[trim={12.1mm 95.3mm 14.9mm 0mm},clip,width=\linewidth]{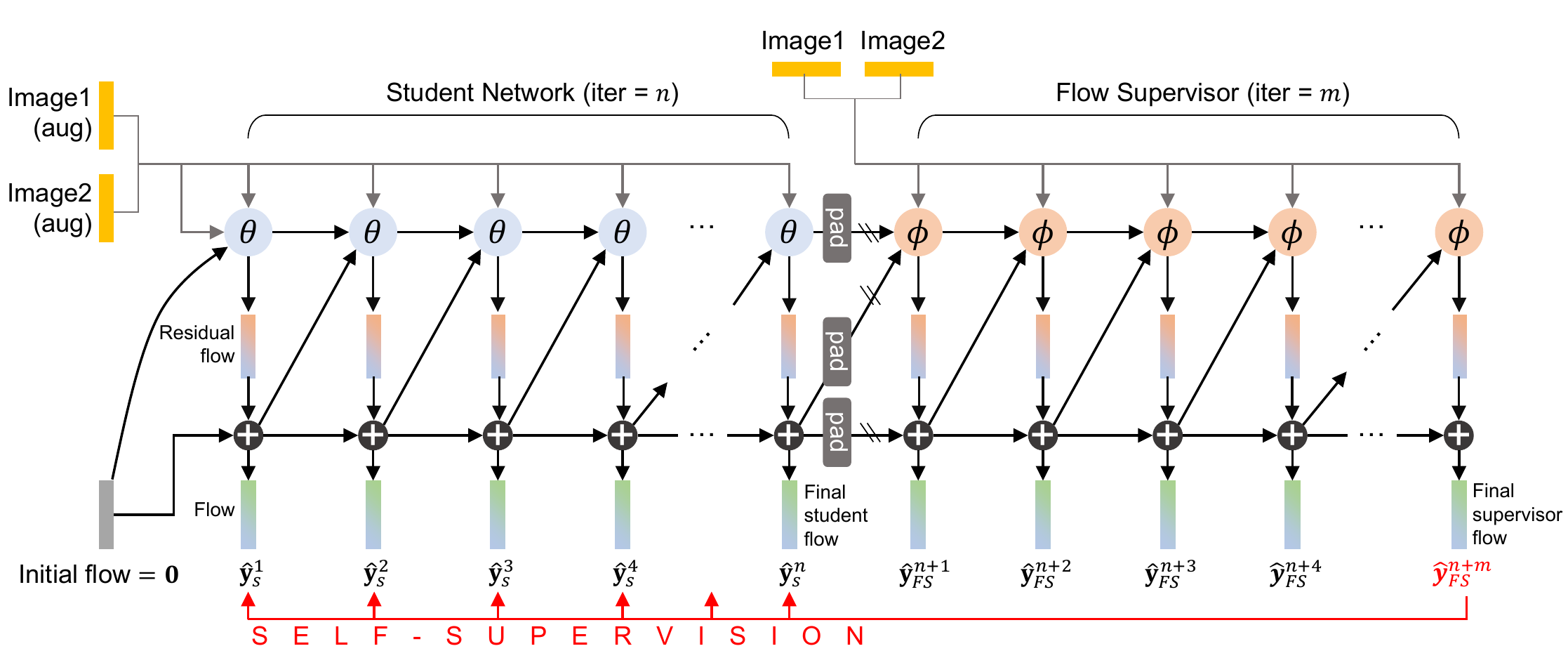}
 \caption{Overall structure for self-supervision}
 \end{subfigure}
\begin{subfigure}[b]{0.4\linewidth}
 \centering
 \includegraphics[trim={12.1mm 21.9mm 123mm 113.6mm},clip,width=\linewidth]{figure/supp_overall.pdf}
 \caption{A refinement step of RAFT~\cite{teed2020raft}}
\end{subfigure}\hfill
\centering
\caption{\textbf{Detailed architecture.} \textbf{(a)} 
summarizes the detailed structure of our flow supervisor. Our flow supervisor shares the design
of iterative refinement RNN module of RAFT. Since we feed full images to the flow supervisor, we pad the outputs of student network to feed them to the supervisor. \textbf{(b)} depicts one refinement step of RAFT with the feature encoder and context encoder. For technical details of each layer, please refer to \cite{teed2020raft}.}
\label{fig:overall_supp}
\end{figure}

We show the implementation of our method in Fig.~\ref{fig:overall_supp}.
Our architecture is based on RAFT~\cite{teed2020raft},
where the iterative refinement plays an important role.
The self-supervision process is similar to the supervised learning,
where the intermediate flows $\hat{\mathbf{y}}^i_s$ are supervised by
a target flow $\mathbf{y}$.
In the RAFT paper, the decaying parameter $0 < \gamma \leq 1$ is used in the loss function:
\begin{equation}
    \ell_\text{sup} = \sum_{i=1}^n \gamma^{n-i} \|\hat{\mathbf{y}}^i_s - \mathbf{y} \|_1.
\end{equation}
Similarly, we apply the decaying strategy in our self-supervision by minimizing:
\begin{equation}
    \ell_\text{FS} = \sum_{i=1}^n \gamma^{n-i} \rho(\hat{\mathbf{y}}^i_s - \hat{\mathbf{y}}^{n+m}_\text{FS}),
\end{equation}
where $\hat{\mathbf{y}}^{n+m}_\text{FS}$ is the pseudo label predicted by the flow supervisor.
Specifically, we use $\gamma=0.8$ ($\ell_\text{sup}$), $\gamma=0.8$ ($\mathcal{L}_\text{FS}$), and $\gamma=1.0$ ($\mathcal{L}_\text{TS}$). 
For KITTI, we use 
$\gamma=0.8$ ($\mathcal{L}_\text{TS}$, $\mathcal{L}_\text{TU}$) for the supervisor model.

\subsection{Padding Operation}
As mentioned in the main text,
we use a cropping operation to give supervision
from the supervisor network to the student network.
Passing student outputs requires the student outputs to be aligned with the 
uncropped images (i.e., teacher inputs).
Since RAFT use $1/8$ processing resolution,
we use a padding operation performed at $1/8$ resolution,
and the random offset coordinates for cropping operation 
is constrained to multiples of 8;
this results in sizes of all inputs -- 
augmented, privileged, and crop offsets --
to be multiples of 8.

\subsection{Optimization}
In the fine tuining stage, we use batch size 1 each from a labeled dataset and an unlabeled dataset, which requires a single RTX3090 GPU, and it takes 
about one day to converge.
We use Adam optimizer~\cite{kingma2014adam}, 
and we decay the learning rate
from $10^{-5}$ by $1/2$ every 25,000 steps.
Different from supervised training, we use the generalized Charbonnier loss $\rho(\cdot)$ in $\ell_\text{sup}$, $\ell_\text{self}$, and $\ell_\text{FS}$ for semi-supervised training. 
Detailed hyper-parameters and reproducible experimental settings
are provided in our code.

\paragraph{Unsupervised Loss.}
In $\mathcal{L}_\text{TU}$ and ablation results, we use
the unsupervised loss $\ell_\text{unsup}(\cdot)$ for comparison.
As mentioned in the main paper, we use the photometric loss, 
occlusion handling, and the smoothness loss,
which are brought from SMURF implementation~\footnote{https://github.com/google-research/google-research/tree/master/smurf}.
Following the code, we use \texttt{census=1}, \texttt{smooth1=2.5},
\texttt{smooth2=0.0}, and \texttt{occlusion='wang'} for Sintel;
\texttt{census=1.0}, \texttt{smooth1=0.0},
\texttt{smooth2=2.0}, and \texttt{occlusion='brox'} for KITTI.
We do not use the self-supervision loss (\texttt{selfsup=0.0}) since our method includes the similar
self-supervision strategy.

\section{Self-Supervision Example}
\begin{figure}[t]
\centering
 \includegraphics[trim={0mm 70.8mm 0mm 0mm},clip,width=\linewidth]{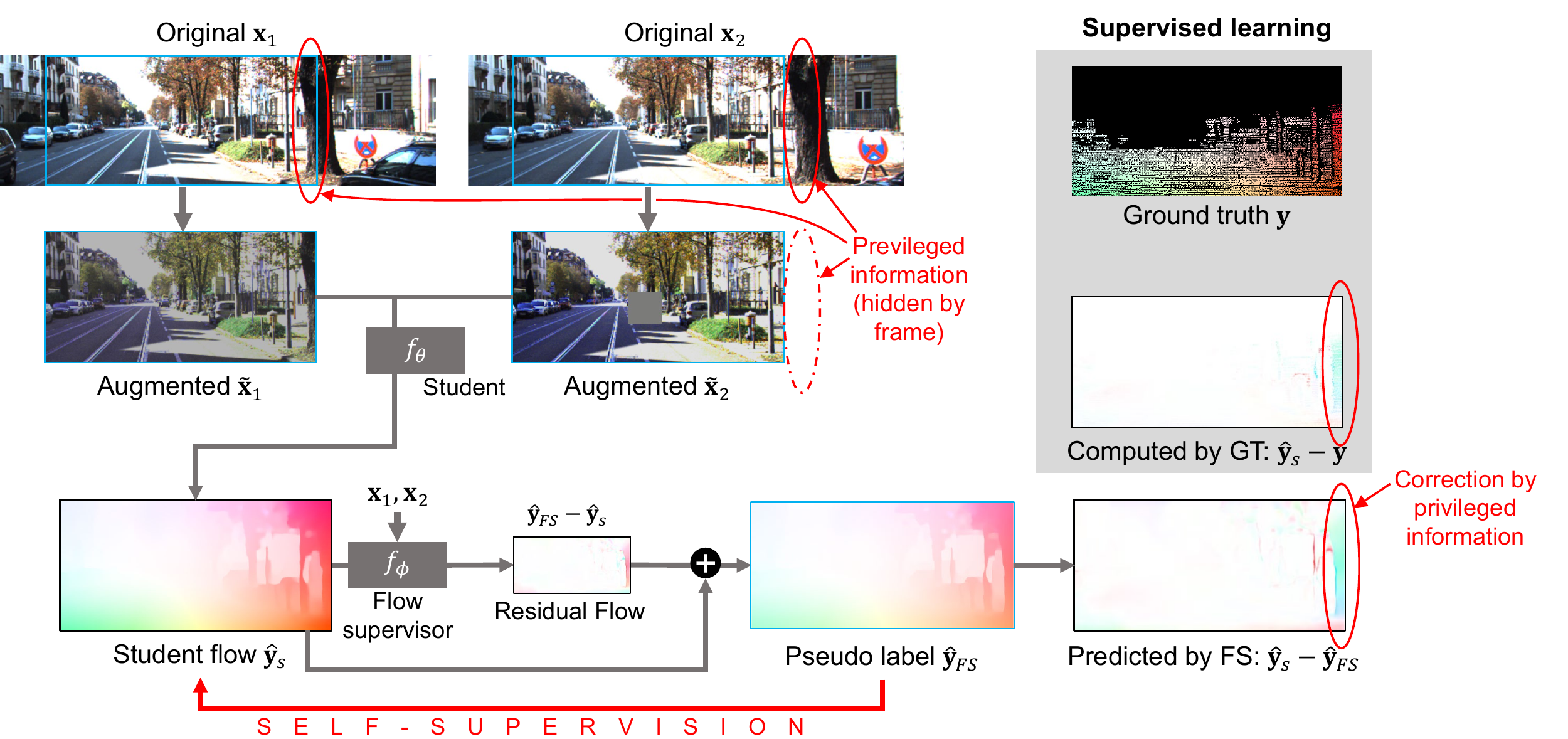}
\caption{\textbf{Self-supervision example.}}
\label{fig:supex_supp}
\end{figure}

Our self-supervision is performed by the flow supervisor
which is conditioned on the student outputs and 
clean inputs.
The process is summarized in Fig.~\ref{fig:supex_supp}.
In the example, we show consecutive frames from a driving scene,
where the observer is moving forward, so that objects near the image frame
are hidden by the frame.
For example, the tree on the right side is not visible in the second cropped image,
while the original second image contains the tree.
Thus, the teacher prediction can correct the student prediction 
by utilizing the privileged information, as shown in the figure.
Compared to the ground truth, we can observe the correcting direction $\hat{\mathbf{y}}_s - \hat{\mathbf{y}}_\text{FS}$ is close to $\hat{\mathbf{y}}_s - \mathbf{y}$ computed by GT.
Thus, our flow supervisor can generate a desirable supervision signal
to guide the student network by the privileged inputs.

\end{document}